\documentclass{article}
\usepackage{spconf,amsmath,graphicx}

\usepackage{framed,multirow}
\usepackage{amssymb}
\usepackage{latexsym}
\usepackage{lipsum}
\usepackage{url,cite}
\usepackage[table]{xcolor}
\usepackage{graphicx}
\usepackage[tight,footnotesize]{subfigure}
\usepackage{dsfont}
\usepackage{tabularx, tabulary, multirow, bigstrut}
\usepackage{booktabs}
\usepackage[american]{babel}
\usepackage[vlined]{algorithm2e} 
\usepackage{fixltx2e}
\usepackage{lipsum}
\usepackage{enumitem}
\usepackage[pagebackref=false,breaklinks=true,letterpaper=true,colorlinks,bookmarks=false]{hyperref}
\graphicspath{ {./figs/} }

\def\x{{\mathbf x}}
\def\y{{\mathbf y}}
\def\z{{\mathbf z}}

\title{AnimGAN: A Spatiotemporally-Conditioned Generative Adversarial Network for Character Animation}
\name{Maryam Sadat Mirzaei$^{1,2}$ \qquad Kourosh Meshgi$^{2}$ \qquad Etienne Frigo$^{3}$ \qquad Toyoaki Nishida$^{1,2}$\thanks{This work was supported by JSPS KAKENHI (No. 19H01120).}
}
\address{$^{1}$ Graduate School of Informatics, Kyoto University, Japan \\
    $^{2}$ RIKEN Center for Advanced Intelligence Project (AIP), Japan \\
    $^{3}$ Engineering School, University of Savoie Mont Blanc, France}

\begin{document}
\setlength{\abovedisplayskip}{3pt}
\setlength{\belowdisplayskip}{3pt}
%
\maketitle
\begin{abstract}
Producing realistic character animations is one of the essential tasks in human-AI interactions. Considered as a sequence of poses of a humanoid, the task can be considered as a sequence generation problem with spatiotemporal smoothness and realism constraints. Additionally, we wish to control the behavior of AI agents by giving them what to do and, more specifically, how to do it. We proposed a spatiotemporally-conditioned GAN that generates a sequence that is similar to a given sequence in terms of semantics and spatiotemporal dynamics. Using LSTM-based generator and graph ConvNet discriminator, this system is trained end-to-end on a large gathered dataset of gestures, expressions, and actions. Experiments showed that compared to traditional conditional GAN, our method creates plausible, realistic, and semantically relevant humanoid animation sequences that match user expectations. 
\end{abstract}
\section{Introduction}
\label{sec:intro}
One of the challenges of developing virtual agents is the scarceness of proper animations and reactions to establish meaningful, realistic, and engaging interaction with the user. Virtual agents require a battery of non-repetitive gestures, expressions, and actions to promote the interaction, which calls for generating customizable actions, big datasets of stored animations, or efficient online generation techniques. Designing and real-time customization of the actions requires a great deal of expertise; thus, significant progress has been made to realize the automatic generation of facial expressions, lip sync \cite{zhou2019talking}, gaze \cite{ruhland2015review} and hand gestures \cite{kucherenko2020gesticulator}. Synthesizing new motion by using autoregressive techniques gained some attention recently. However, existing methods tend to freeze or diverge
after a couple of seconds due to an accumulation of errors in the closed training loop of the networks\cite{zhou2018auto}. 

Generative adversarial networks (GANs), consists of a generator and a discriminator networks that are trained adversarially to generate new data with the same statistics as the training set, e.g., by generating new photographs that look at least superficially authentic to human observers, having many realistic characteristics \cite{karras2019analyzing}. 
One of the practical problems of GANs is multi-modal data. Despite the advancements in this domain, it is still challenging to scale such models to accommodate an extremely large number of predicted output categories. One of the solutions is to provide the two networks with extra auxiliary information in the form of class labels or data from other modalities. Class-conditional GANs generates much better samples than GANs that were free to generate from any class \cite{chen2016infogan}. Additionally, even if class information does not explicitly fed to the generator, the sample quality may improve  \cite{salimans2016improved}, and training the discriminator to recognize specific classes of real objects is sufficient.

\begin{figure}[!t]
\centering
\includegraphics[width=1\linewidth]{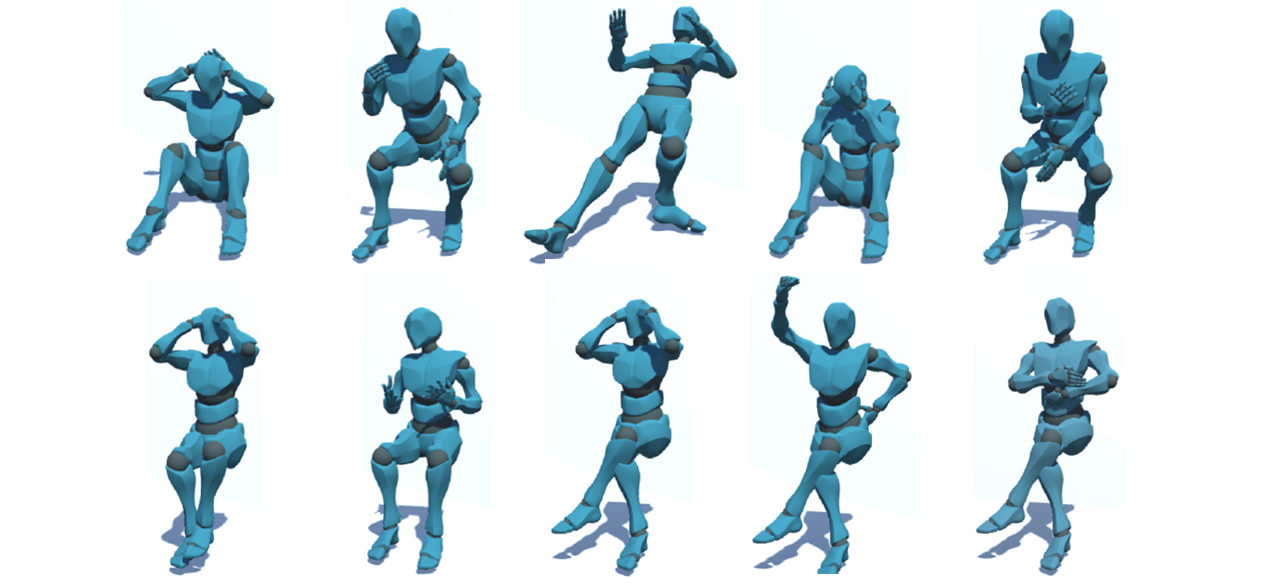}

\caption{The C-GAN \textbf{(first row)} is presented with one of the available action labels \texttt{\small sitting-talking-angrily} while the input sequence to our proposed STC-GAN \textbf{(second row)} depicts a sitting person who is angrily talking with his hands. The sequence generated with the former tends to covers a wide variety of poses, some semantically different (e.g., relaxed, relaxed, etc.) and some aesthetically different (e.g., sitting on the floor), whereas the sequences generated with proposed method tend to have similar semantics and spatiotemporal dynamics with the input while maintaining diversity. Visualized using Unity3D, character taken from Mixamo(c).}
\label{fig:concept}
\vspace{-0.5 cm}
\end{figure}
%

To condition the generation process on the input, Conditional GAN is proposed in \cite{mirza2014conditional}, which provide both generator and discriminator with additional data about the part of distribution that the generator is expected to model, for example, having a \texttt{\small sitting-talking} sequence as an input, tell both generator and discriminator that what part of \textit{action space} is considered. In \cite{mirza2014conditional} authors introduced the data label as the auxiliary input, which only defines the region-of-interest in action space. However, their method is unable to condition the output with a specific fine-grained semantic and spatiotemporal dynamics of the input since the labels are by nature impose coarse clustering in the action space (Figure \ref{fig:concept}). To this end, we propose to input an instance of the expected input as a sample of the expected fine-grained region of the input distribution. Specifically, we propose to use spatiotemporally-conditioned GAN (STC-GAN) to generate actions semantically and structurally close to the input. In this study, we
\begin{itemize}[noitemsep,leftmargin=*]
\item proposed STC-GAN for fine-tuned semantically and syntactically consistent sequence generation;
\item created animGAN that employs the proposed framework for character animation;
\item gathered and unified a large dataset for realistic character motions, including gestures, expressions, and actions.

\end{itemize}

\section{Background}
\label{sec:bkg}
\noindent\textbf{Animation Generation:}  Considering the cost of gathering mocap data, animation generation has been studied for different applications such as games, movie making and human-computer interaction \cite{kovar2008motion,ramachandran2016unsupervised}. With the prevalence of deep networks different lines of research has emerged in this domain e.g., generating gait dynamics and foot kinematics \cite{holden2017phase}, sequence generation given early few frames \cite{ghosh2017learning,zhou2018auto}, from textual description \cite{lin2018generating,plappert2018learning}, physics-based motion controllers \cite{kwon2017momentum}, and reinforcement learning for mimicing human \cite{peng2018sfv}.   

\noindent\textbf{GANs for Sequence Generation:}
Modeling the sequence of data is usually done using Recurrent neural networks (RNNs), where they are used to trained to predict the next stage of the sequence, and were initially used along with adversarial training in RNN-GAN \cite{mogren2016c} to model the whole joint probability of a sequence and be able to generate the sequence of data. 
A conditional LSTM-GAN is proposed in \cite{yu2019conditional} to generate a novelty music sequence, given the lyrics as additional context to instruct these two networks. Generally, GANs give score/loss for the entire generated sequence when it has been generated, and for a partially generated one, it is non-trivial to balance the quality of generated part versus the part to be generated. To this end, the data generator is modeled as a stochastic policy in reinforcement learning \cite{yu2017seqgan} via Monte Carlo tree search, a reward at every generation step \cite{li2017adversarial}, or getting feedback from the discriminator \cite{tuan2019improving}.

\section{Model}
\label{sec:method}
To realize AnimGAN, a version of the proposed STC-GAN that generates character animations that are semantically and spatiotemporally consistent with the input sequence, we propose to generate a set of consecutive motions of the skeleton of a human body from an input sequence of actions, centered on human in a feed-forward manner. During training, we assume that all images are annotated with 2D joints. We also prepared a dataset of annotated video sequences of human bodies doing different actions, where the data annotations come from ground truth, 2D pose estimation systems, or projecting 3D joints from MoCap data into an image plan. 
We proposed a spatiotemporally-conditioned GAN to preserve the temporal semantics while generating novel realistic sequences similar to the initial one. A sequence of pose embedding vectors (one from each frame) concatenated with noise vectors is taken as the input of the generator network. The sequence of consecutively generated poses is then passed to the discriminator that tries to distinguish them from the real-world actions present in the action dataset. In addition, the generator is equipped with a temporal smoothness regularization to cancel out the jitter in the joint localization in the dataset and input data and to shrink the generation space to yield more plausible results while speeding up the training.

\subsection{Problem Formulation}
Proposed by Goodfellow et al. \cite{goodfellow2014generative}, GAN simultaneously train two competitive networks, a generator $\mathcal{G}$ and a descriminator $\mathcal{D}$ with opposing objectives. $\mathcal{G}$ tries to capture data distribution of the training set, whereas $\mathcal{D}$ tries to distinguish samples produced by the generator from the real ones. In original GAN, generator takes a Gaussian noise vector $\z$, and converts it to an synthetic sample $\tilde{\y} = \mathcal{G}(\x)$. Playing a zero-sum game, the GAN is improving both $\mathcal{G}$ and $\mathcal{D}$ with the following loss:
\begin{equation}
\label{eq:gan}
\begin{aligned}
\min _{\mathcal{G}} \max _{\mathcal{D}} V(\mathcal{D}, \mathcal{G}) &=\mathbb{E}_{\mathbf{y} \sim p_{\text {data }}(\mathbf{y})}[\log \mathcal{D}(\mathbf{y})] \\
&+\mathbb{E}_{\mathbf{z} \sim p_{\mathbf{z}}(\mathbf{z})}[\log \big(1-\mathcal{D}(\mathcal{G}(\mathbf{z}))\big)]
\end{aligned}
\end{equation}

In the conditional GAN, both $\mathcal{G}$ and $\mathcal{D}$ are provided with the auxiliary input $\x$, where generator $\mathcal{G(\z|\x)}$ and discriminator $\mathcal{D(\z|\x)}$ plays the min-max game in Equation \eqref{eq:gan}. 

Taking a sequence $\mathcal{X}=\left(\x^{1}, \cdots, \x^{|\mathcal{X}|}\right)$ as input, our goal is to generate a sequence $\mathcal{Y}=\left(\y^{1}, \cdots, \y^{|\mathcal{Y}|}\right)$ that is semantically similar to $\mathcal{X}$, preserve the speed of the action in $\mathcal{X}$, and is realistic. Inspired by \cite{yu2019conditional}, we defined the loss functions ($m$ is mini-batch size) as 
\begin{equation}
L_{\mathcal{G}}= \sum_{i=1}^{m} \log \left(1-\mathcal{D}(\mathcal{G}\left(\z^i | \x^i)\right)\right) + L_{\mathrm{\mathcal{ST}}}
\end{equation}
\begin{equation}
L_{\mathcal{D}}=- \sum_{i=1}^{m}\left[\log \mathcal{D}\left(\y^i | \x^i\right) +\log \left(1-\mathcal{D}\left(\mathcal{G}\left(\z^i | \x^i\right)\right)\right)\right]
\end{equation}
where $L_{\mathrm{\mathcal{ST}}}$ is the spatiotemporal conditioning loss.

\subsection{Spatiotemporal Conditioning}
For each specific action, a subset of joints is activated, and for different actions, a different set of joints are used. Hence, in action classification, only the spatial joints that are spatially activated should be focused (hereafter called \textit{main joints}), and the rest could be ignored as they are mostly non-discriminant, occluded, or noisy \cite{weng2017spatio}.
Using spatiotemporal conditioning term serves two main purposes: \textit{(i)} matching the generated pose to its previous pose to avoid large displacement and drastic velocity changes in the main joints for the actions, \textit{(ii)} enforcing spatiotemporal similarity between main joints generated sequence of pose and the corresponding pose from input sequence, imposing smoothness constraint of the sequence. Here, we proposed the following loss function.
\begin{equation}
L_{\mathrm{\mathcal{ST}}} = \sum_{i=1}^{m} \lambda_1 \phi(\x^i,\tilde{\y}^i) +\lambda_2 \phi(\tilde{\y}^{i-1},\tilde{\y}^i) + \epsilon
\end{equation}
in which $\lambda_1$ and $\lambda_2$ are importance factors of conditioning and smoothness terms, $\epsilon$ is a small positive constant, and $\phi(.)$ is the mean-squared distance of the spatial location and velocity of the main joins. Here, the main joints are detected using spatiotemporal naive-bayes nearest neighbor (ST-NBNN \cite{weng2017spatio}) of the input sequence $\x$ that uses a bilinear classifier to learn spatio-temporal weights for important spatial joints and temporal stages in the framework of NBNN \cite{boiman2008defense}.

\subsection{Architecture}
\begin{figure}[!h]
\centering
\includegraphics[width=1\linewidth]{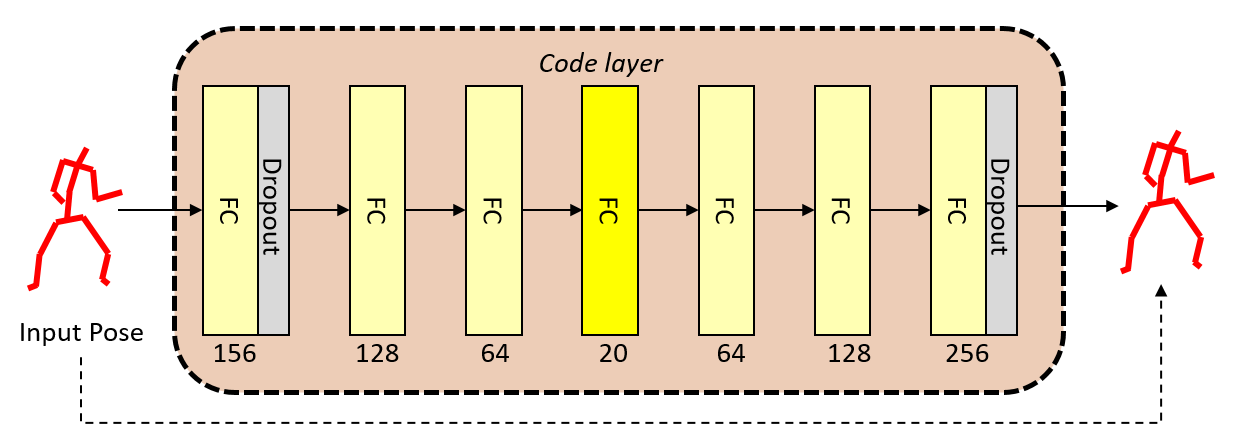}
\caption{Sparse autoencoder used for pose embedding}
\label{fig:ae}
\vspace{-0.5 cm}
\end{figure}

\noindent\textbf{Pose Embedding:} To handle the large imbalance in the number of the videos for different action categories, Lin et al.\cite{lin2018generating} employed a neural autoencoder to learn a compact representation of human motions. Here, the pose in each frame is defined by the joint rotation matrices of the BVH (BioVision Hierarchy) representation of the animation sequence in Quaternion format ($\mathbb{R}^{21\times4}$ for each pose). We trained a L1-sparse autoencoder\cite{ng2011sparse} on the pose data, to reduce the dimensionality while capturing some spatial relationship in the represented skeleton (Figure \ref{fig:ae}).

\begin{figure}[!h]
\centering
\includegraphics[width=1\linewidth]{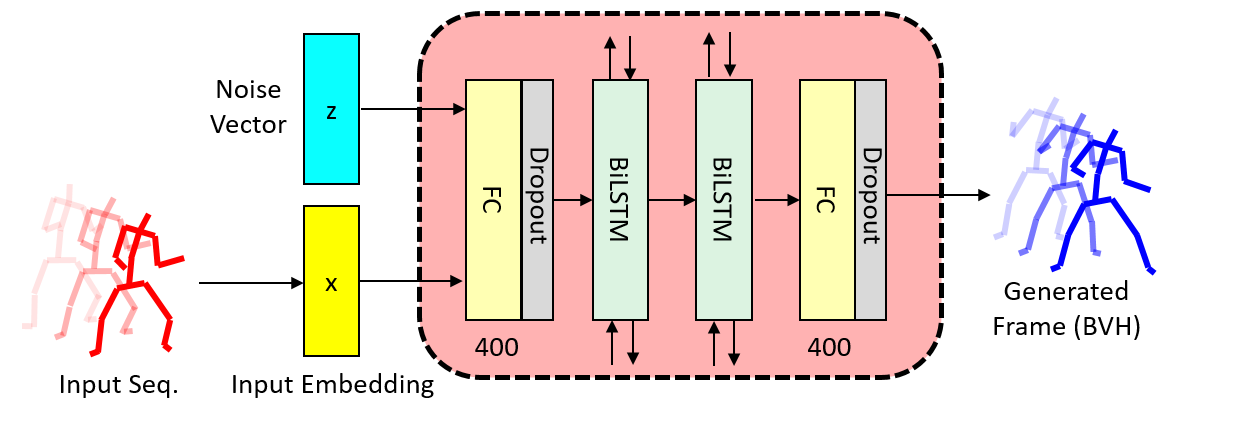}
\caption{Generator architecture to generate joints rotation matrices from input pose embeddings $\x$ and random noise $\z$.}
\label{fig:g}
\vspace{-0.5 cm}
\end{figure}

\noindent\textbf{Generator:}
The generator is to learn the distribution of input videos from human actions, generate realistic samples, and fool a real/fake discriminator in doing so. In this work, each action sequence has $k$ frames, which need $k$ LSTM cells to learn the strong semantic and loose spatiotemporal alignment between input and output sequence. The first layer in the generator network maps the 50D input (input pose embedding $\x \in \mathbb{R}^{20}$ concatenated with Gaussian noise vector $\z \in \mathbb{R}^{30}$) to the $k$D LSTM cells. After the second BiLSTM layer, another fully connected layer (+ dropout) converts the cell outputs to $\mathbb{R}^{21\times4}$ (Fig. \ref{fig:g}) to be reshaped into a BVH animation.

\begin{figure}[!h]
\centering
\includegraphics[width=1\linewidth]{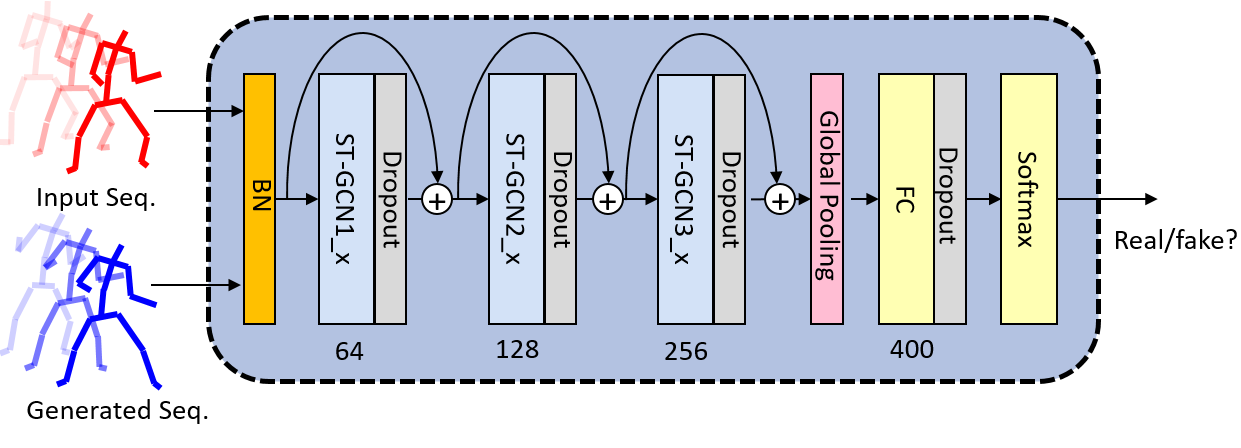}
\caption{Discriminator architecture, conditioned on input pose embeddings $\x$ and detects if the pose $\tilde{y}$ is real or fake. The network consists of 9 ST-GCN layers, each followed by a dropout layer, and pooled every 3 layers. The residual connections is applied to each convolution block. 
}
\label{fig:d}
\vspace{-0.5 cm}
\end{figure}

\noindent\textbf{Discriminator:}
To determine whether a sequence of poses comes from real human actions or from the generator, we used a spatiotemporal graph convolution network (ST-GCN) architecture, as proposed by \cite{yan2018spatial}. In this GCN, the sequence of poses is represented with a graph in which all nodes of the pose in time $t$ are connected with \textit{intra-body} edges based on the body hierarchy proposed in BVH format. Corresponding joints in immediate times frames $t-1$ and $t+1$ are also connected via \textit{inter-frame} edges. The input of the ST-GCN is the joint coordinates w.r.t. the hip position, calculated based on the rotation matrices and the standard character limb lengths (the distance between two joints) defined for each frame in the BVH animation abstraction format. A global pooling (to handle the input sequences with indefinite length) is then performed on the ST-GCN's output tensor followed by two dense layers to determine if a sequence is from real action data or synthesized by the generator (Figure \ref{fig:d}).

\subsection{Dataset} 
To create a rich database to train pose embedding, action generator, and the discriminator, we gathered real action sequences in the form of videos and motion capture data. The data are downsampled to 5 fps, and the results are converted into 21-joint BVH format. Long sequences are trimmed to a maximum of 300 frames. We gather video sequences from different publicly available datasets:
\begin{itemize}[noitemsep,leftmargin=*]
\item Emotional Body Motion database\cite{volkova2014mpi} that focuses on emotion expression in human-human interaction using MoCap;
\item NTU-RGB+D 120 \cite{liu2019ntu} that is captured with 3 Kinect V2 concurrently and comes with 3D skeletal data;
\item Kinetics Dataset \cite{kay2017kinetics} that is a high-quality dataset of YouTube video URLs which include a diverse range of focused human actions for which the 3D estimation of the joint locations are estimated using HMR \cite{kanazawa2018end};
\item KIT Motion Language Dataset \cite{plappert2016kit} aiming to capture semantic representations of human activities for which the master motor map (MMM) data \cite{terlemez2014master} is converted to BVH;
\end{itemize}
The gathered dataset includes a range of communication gestures and signals, behaviors, and emotions expressed through body motion, actions (daily actions, sports, medical conditions) and interactions (with humans, environment, objects). 

\subsection{Implementation Details}
\noindent\textbf{Training STC-GAN:}
In our implementation, we first train the pose embedding using 1M randomly selected poses from the database sequences, using batch gradient descent and learning rate of 10$^{-4}$ and dropout rate of 0.5.
The discriminator is trained using stochastic gradient descent, with a learning rate of 0.01 decayed by 10\% every ten epochs. The generator is trained with Adam optimizer, with a learning rate of 0.1 that decays linearly with the progress of the training and momentum of 0.9. The dropout probability in both $\mathcal{G}$ and $\mathcal{D}$ is 0.5. Additionally, the discriminator is trained with noisy labels in accordance with \cite{salimans2016improved}. The activation function in fully-connected layers are LeakyReLU (as suggested in \cite{shi2016real}). We set $\lambda_1,\lambda_2 \in [0,1]$ by cross-validation.

\begin{figure}[!h]
\centering
\includegraphics[width=0.49\linewidth]{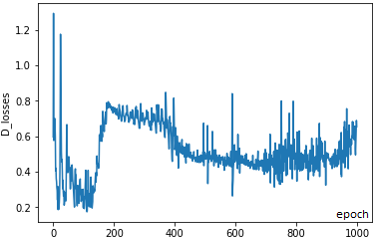}
\includegraphics[width=0.49\linewidth]{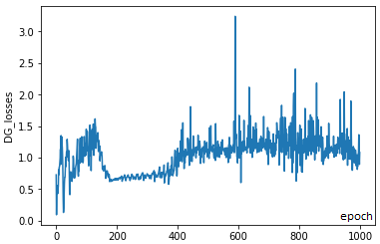}
\caption{Early in the training, the generator's error surpass the spatiotemporal regularization term, but after that the equilibrium between generator and discriminator starts to emerge.}
\label{fig:training}
\vspace{-0.5 cm}
\end{figure}

\noindent\textbf{Data Augmentation:}
The input data is in the form of Quaternion rotation vectors for each joint. The dataset in this format is easy to be manipulated and augmented with new variations ofit is observed that sample quality improves existing sequences. To create a balanced dataset for training, we cluster the existing sequences using the pose embedding, and select the most underrepresented cluster and try to modify the sequence using one of the following operators to generate a new sequence: 
\textit{(i) Mutating} (additive noise), 
\textit{(ii) Cross-over} of the upper and lower body segments of two sequences, 
\textit{(iii) Halving} the join rotation span, 
\textit{(iv) Mirroring}.
The number of clusters is gradually incremented to enable the data augmentation to bootstrap fin-tune under-represented actions.

\noindent\textbf{Synthetic Hard Negatives:}
We create unrealistic or implausible action sequences or body poses and provide them to $\mathcal{D}$ to improve its accuracy. To this end, we apply one of the following operators on several sampled sequences from the dataset: 
\textit{(i) Joint rotation} reversal, 
\textit{(ii) Large random additive joint rotation noise}, 
\textit{(iii) halfway bouncing} of the sequence.

\section{Experimental Results}
\label{sec:eval}
\begin{figure}
\centering
\includegraphics[width=0.48\linewidth]{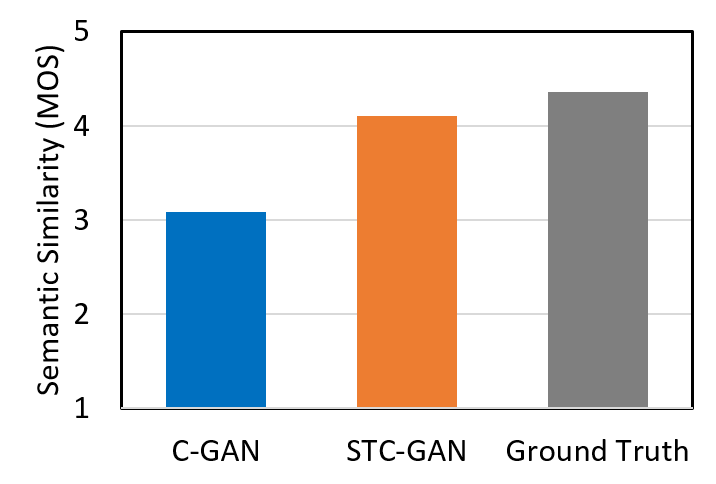}
\includegraphics[width=0.48\linewidth]{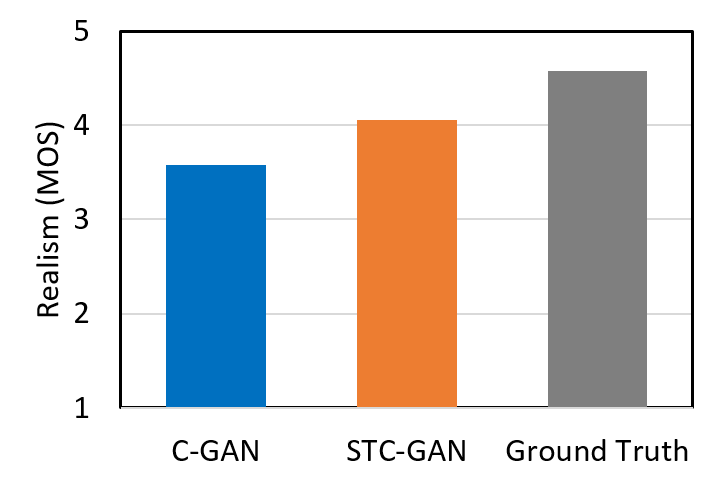}
\caption{Subjective evaluation via mean opinion score (MOS)}
\label{fig:qual}
\vspace{-0.5 cm}
\end{figure}

Here, the experiment setups, validation method, and evaluation results are introduced to investigate the performance of the proposed STC-GAN for character animation generation. 
In this experiment, we show 50 randomly selected animation from the ground truth dataset as the reference sequence. We feed each of these sequences to STC-GAN to generate a novel sequence. We also feed the ground truth's label to C-GAN to generate a competitor sequence. Finally, another video from the dataset with the same label as the reference sequence is retrieved. Thirty participants were asked to rate the semantic similarity of these three sequences (from the database, from C-GAN, and from STC-GAN) with 5-point Likert-scale feedbacks without knowing the source of the sequence. They were also asked to rate the quality (realism) of the three sequences in the same way. Figure \ref{fig:qual} shows that, on average, participants find sequences of STC-GAN significantly more similar to the reference animation. The results also revealed that the generated sequences of the STC-GAN were acceptable in terms of quality. 

\section{Conclusion}
\label{sec:conclusion}
We present an end-to-end GAN for generating human motion animations cued by an input animation. We employed sparse autoencoders to provide pose embedding for the input sequence and developed an LSTM-based generator, a graph convolutional network-based discriminator, a spatiotemporal conditioning loss function based on the changes in spatial and velocity of main joint (detected by ST-NBNN), and we trained the system on our large-scale collected datasets with unified 3D joint annotations in BVH format. In the future, we plan to explore physically-based controller approaches to generate more controllable animations.

\vfill
\pagebreak

\bibliographystyle{IEEEbib}
\small
\def\IEEEbibitemsep{-9pt plus 1pt}
\bibliography{refs}

\end{document}